\documentclass[letterpaper, 10 pt, conference]{ieeeconf}

\IEEEoverridecommandlockouts                              

\usepackage{amsmath,amsfonts}
\usepackage{bm} 
\usepackage{algorithm}
\usepackage[noend]{algpseudocode}
\usepackage{array}
\usepackage[caption=false,font=normalsize,labelfont=sf,textfont=sf]{subfig}
\usepackage{textcomp}
\usepackage{stfloats}
\usepackage{url}
\usepackage{verbatim}
\usepackage{graphicx}
\usepackage{cite}
\usepackage{color}
\usepackage{booktabs}
\usepackage{threeparttable}
\usepackage{nameref}
\usepackage{hyperref}
\usepackage{siunitx}
\usepackage{tabularx}
\usepackage{listings}
\usepackage{xcolor}
\usepackage{fancyhdr}
\usepackage{ragged2e}

\definecolor{codegreen}{rgb}{0,0.6,0}
\definecolor{codegray}{rgb}{0.5,0.5,0.5}
\definecolor{codepurple}{rgb}{0.58,0,0.82}
\definecolor{backcolour}{rgb}{0.95,0.95,0.92}

\title{\LARGE \bf
T\v{r}iVis: Versatile, Reliable, and High-Performance Tool \\ for Computing Visibility in Polygonal Environments
}

\DeclareMathOperator*{\argmin}{arg\,min}

\author{
    Jan Mikula$^{1,2}$, Miroslav Kulich$^{1}$, and Libor P\v{r}eu\v{c}il$^{1}$%
    \thanks{
        This work was co-funded by the European Union under the project Robotics and advanced industrial production (reg. no. \texttt{\footnotesize CZ.02.01.01/\allowbreak{}00/\allowbreak{}22\_008/\allowbreak{}0004590}) and by the Grant Agency of the Czech Technical University in Prague, grant no. \texttt{\footnotesize SGS23/\allowbreak{}175/\allowbreak{}OHK3/\allowbreak{}3T/\allowbreak{}13}.
    }%
    \thanks{
        $^{1}$All authors are with the Czech Institute of Informatics, Robotics and Cybernetics, Czech Technical University in Prague, Jugoslavskych partyzanu 1580/3, Prague~6, 160\,00, Czech Republic.
        \texttt{\footnotesize \{jan.mikula,\allowbreak{} miroslav.kulich, \allowbreak{}libor.preucil\}@cvut.cz}
    }%
    \thanks{
        $^{2}$Jan Mikula is also with the Department of Cybernetics, Faculty of Electrical Engineering, Czech Technical University in Prague, Karlovo namesti 293/13, Prague~2, 121\,35, Czech Republic.
    }%
}

\begin{document}

    \maketitle
    \thispagestyle{fancy}
    \fancyhf{}  
    \renewcommand{\headrulewidth}{0pt}  
    \fancyfoot[C]{\justifying \footnotesize \textcopyright{} 2024 IEEE. Personal use of this material is permitted. Permission from IEEE must be obtained for all other uses, in any current or future media, including reprinting/republishing this material for advertising or promotional purposes, creating new collective works, for resale or redistribution to servers or lists, or reuse of any copyrighted component of this work in other works.}
    \pagestyle{empty}

    \begin{abstract}
        Visibility is a fundamental concept in computational geometry, with numerous applications in surveillance, robotics, and games.
        This software paper presents T\v{r}iVis, a C++ library developed by the authors for computing numerous visibility-related queries in highly complex polygonal environments.
        Adapting the triangular expansion algorithm, T\v{r}iVis stands out as a versatile, high-performance, more reliable and easy-to-use alternative to current solutions that is also free of heavy dependencies.
        Through evaluation on a challenging dataset, T\v{r}iVis has been benchmarked against existing visibility libraries.
        The results demonstrate that T\v{r}iVis outperforms the competing solutions by at least an order of magnitude in query times, while exhibiting more reliable runtime behavior.
        T\v{r}iVis is freely available for private, research, and institutional use at \url{https://github.com/janmikulacz/trivis}.
    \end{abstract}

    \section{Introduction}
    \label{sec:introduction}

    This software paper presents T\v{r}iVis, a C++ library developed by the authors for computing visibility in 2D polygonal environments.
    The authors, focusing primarily on visibility-based route planning solutions for autonomous mobile robots, have encountered severe limitations in the currently available visibility implementations, including the 2D Visibility package in the widely used Computational Geometry Algorithms Library (CGAL).
    Motivated by the need for a fast, robust, versatile, and lightweight solution, the authors developed T\v{r}iVis to address these limitations and provide a high-performance, reliable, and easy-to-use solution for the robotics community and beyond.

    The library is suitable for various applications that involve 2D polygonal environments where frequent visibility queries are essential.
    These applications include the design of security and surveillance systems~\cite{Agarwal2009} and robotics.
    In the field of robotics, specific examples include route planning for efficient environment inspection or search~\cite{Mikula2022,Kulich2022}, multi-agent systems~\cite{Tokekar2015,Maini2021}, and pursuit-evasion games~\cite{Lozano2022,Fletcher2023}.

    This paper is organized as follows:
    Sec.~\ref{sec:definitions} defines the fundamental concepts of visibility and related terms.
    Sec.~\ref{sec:algorithms} provides a literature background on visibility algorithms and describes the primary algorithm used in T\v{r}iVis: the triangular expansion algorithm~\cite{Bungiu2014,Xu2015}.
    This section also details the T\v{r}iVis-specific adaptations of the algorithm for handling other visibility queries.
    Sec.~\ref{sec:design} details the design of T\v{r}iVis, including the library's robustness strategies, internal dependencies, and a short snippet of the library's usage.
    Sec.~\ref{sec:performance} evaluates the performance of T\v{r}iVis on a dataset of complex polygonal environments and compares it with other visibility libraries.
    Finally, Sec.~\ref{sec:conclusion} concludes the paper.

    \section{Definition of Visibility and Related Terms}
    \label{sec:definitions}

\begin{figure}
        \centering
        \subfloat{\includegraphics[width=0.328\columnwidth]{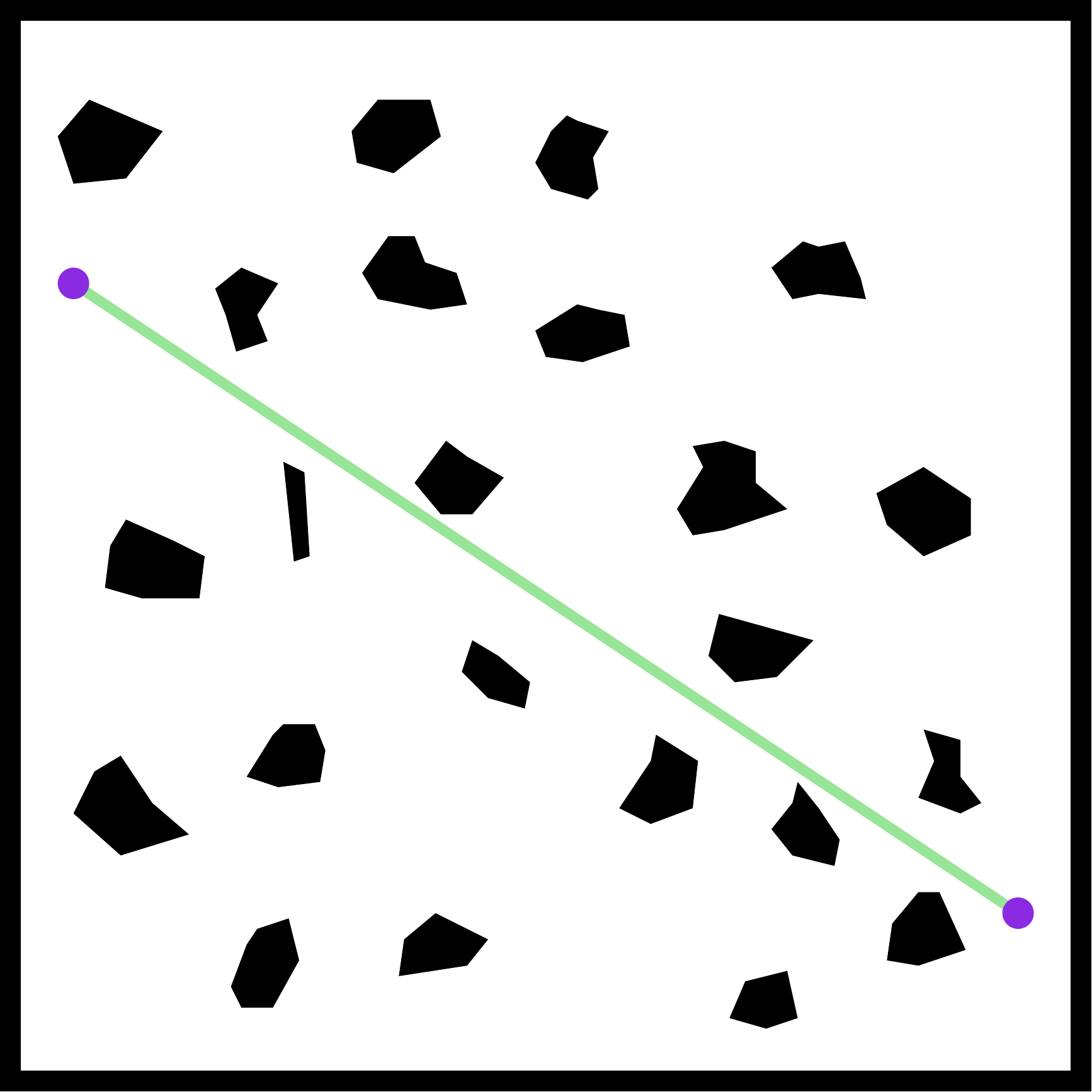}}
        \hfill
        \subfloat{\includegraphics[width=0.328\columnwidth]{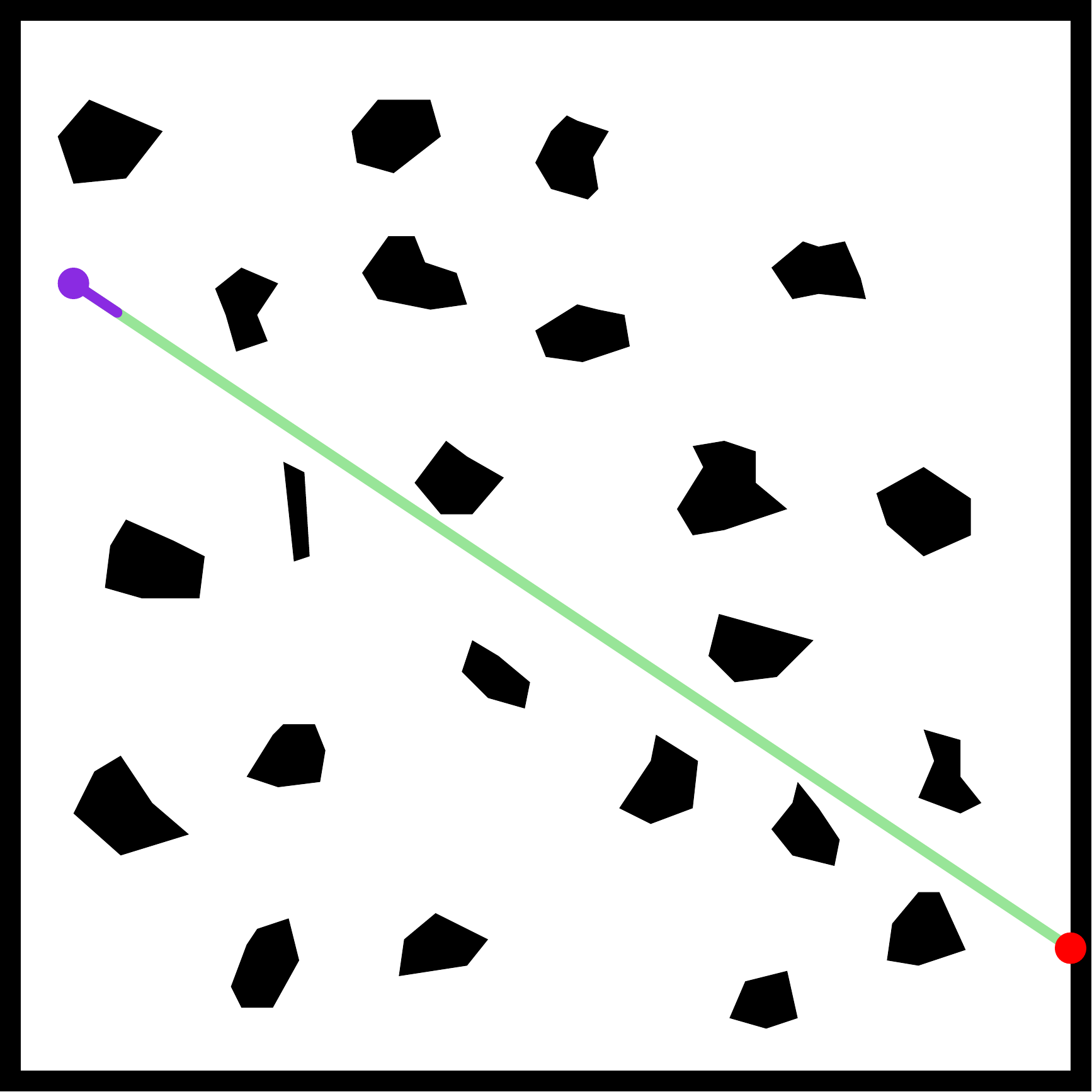}}
        \hfill
        \subfloat{\includegraphics[width=0.328\columnwidth]{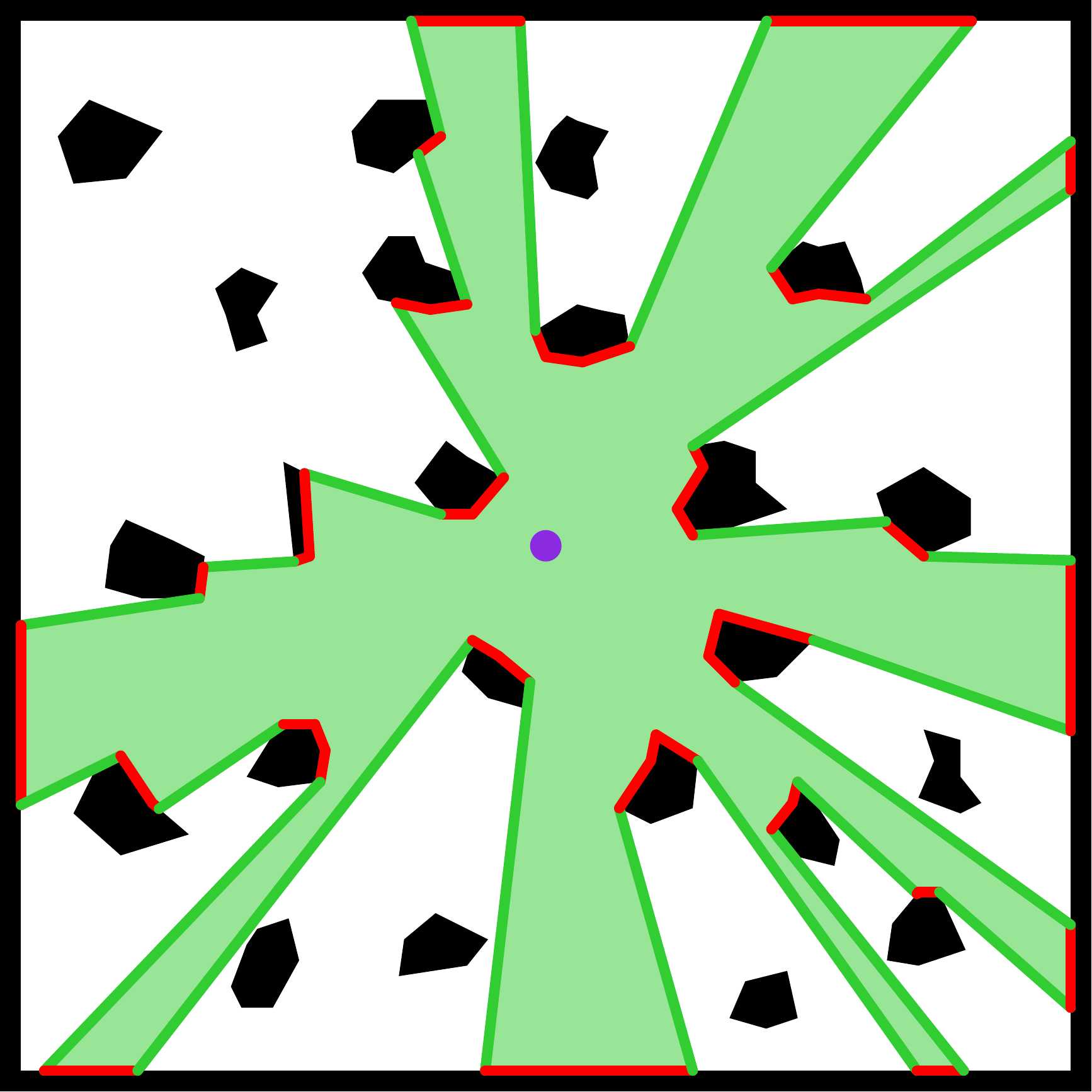}}
        \\\vspace{-0.8em}
        \subfloat{\includegraphics[width=0.328\columnwidth]{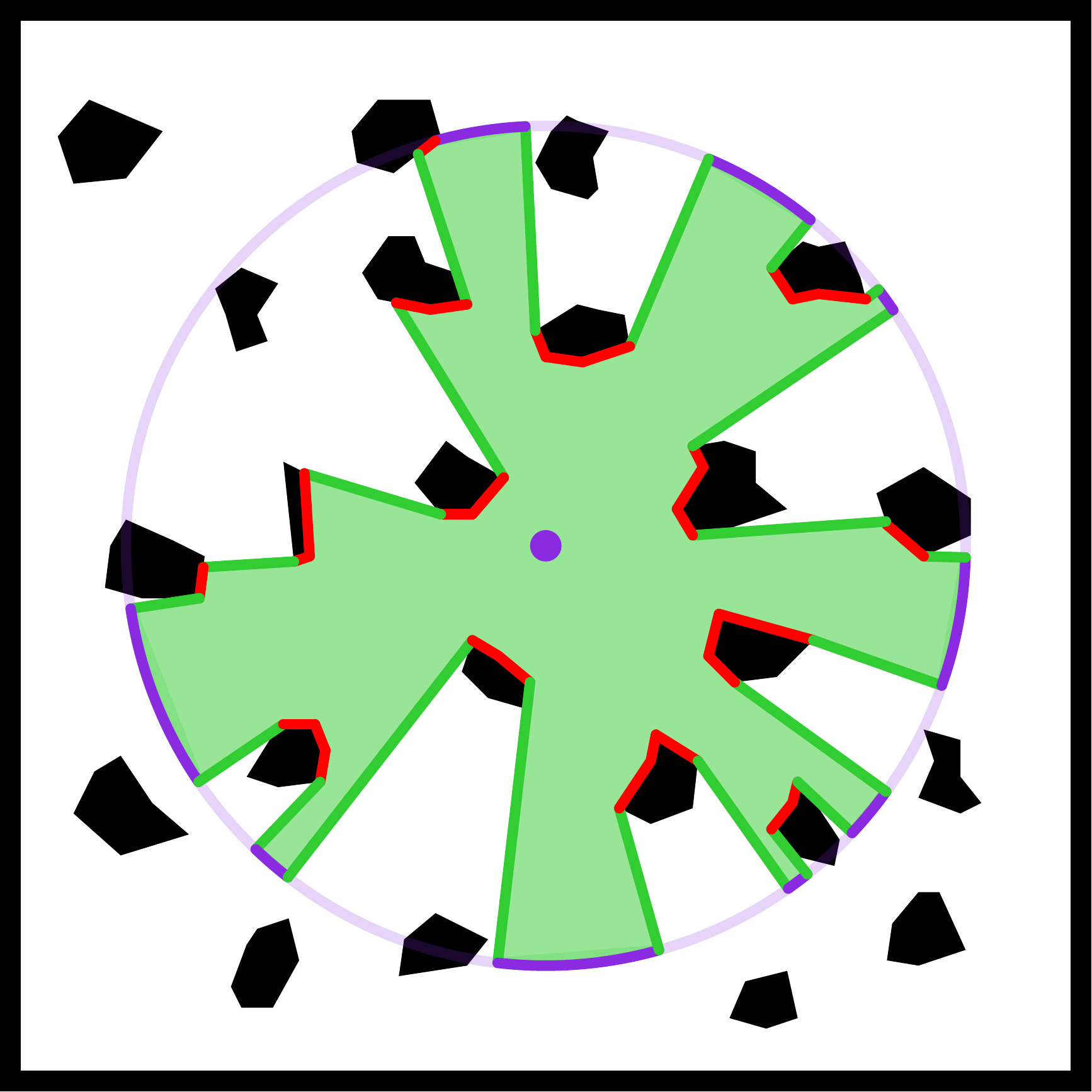}}
        \hfill
        \subfloat{\includegraphics[width=0.328\columnwidth]{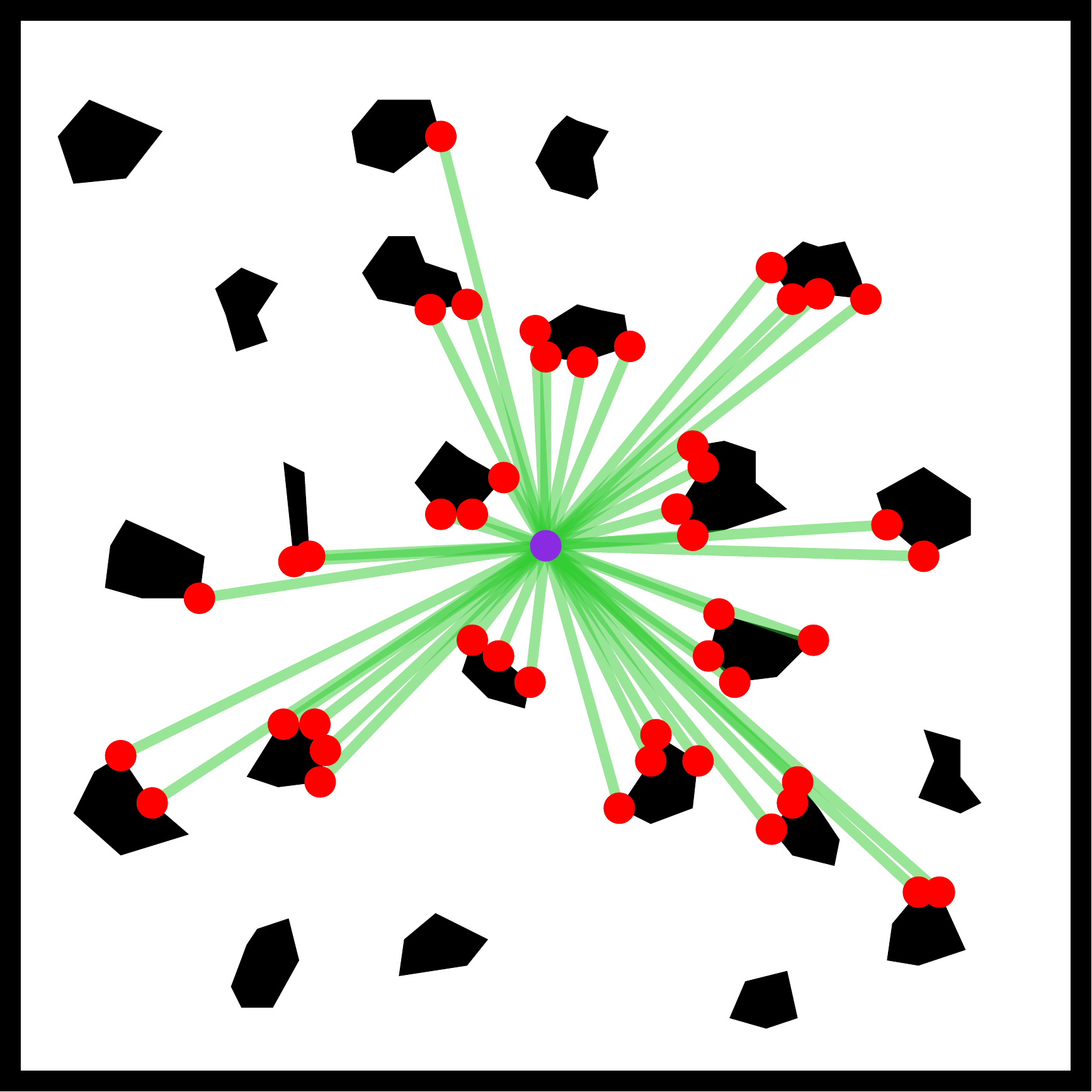}}
        \hfill
        \subfloat{\includegraphics[width=0.328\columnwidth]{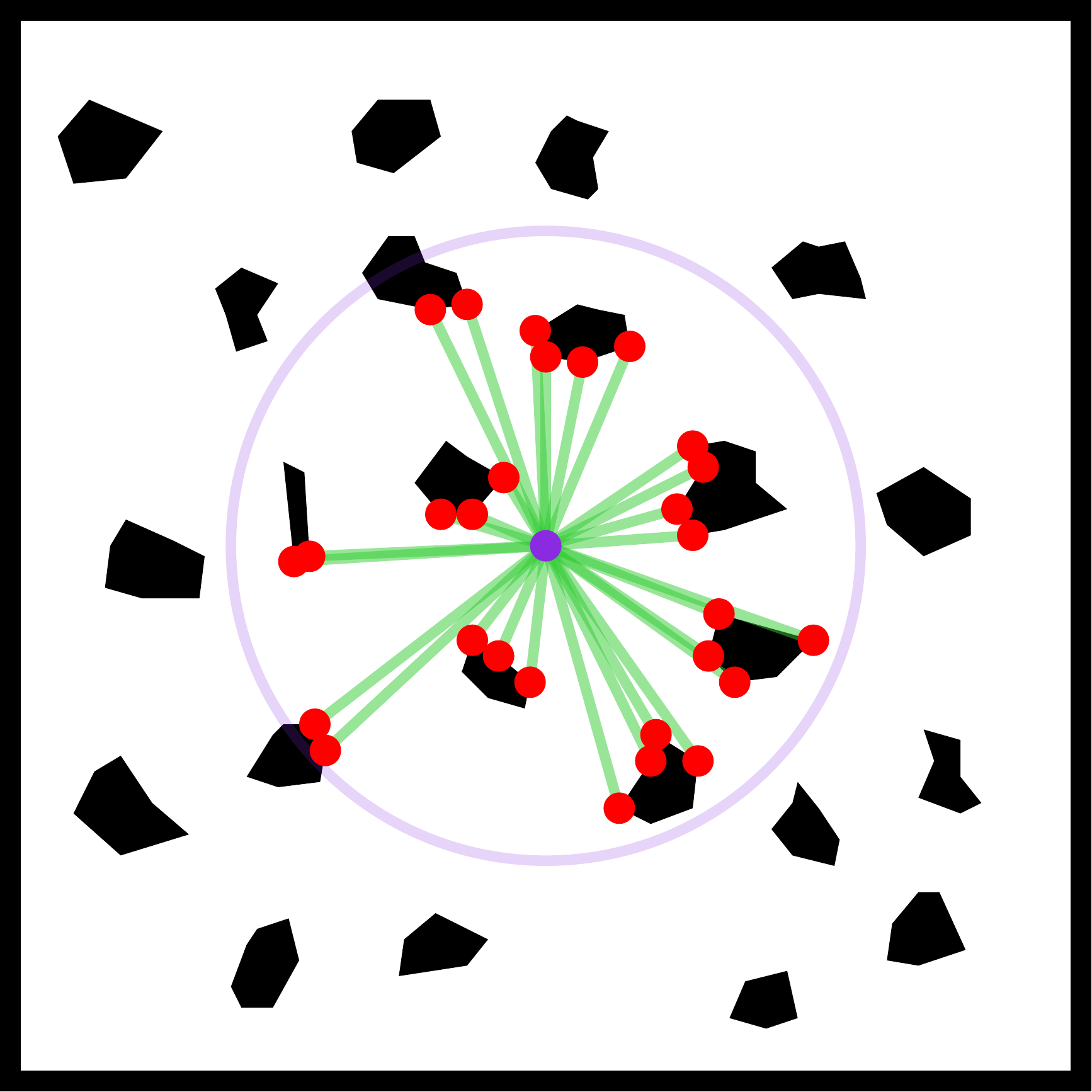}}
        \\\vspace{-0.8em}
        \subfloat{\includegraphics[width=0.328\columnwidth]{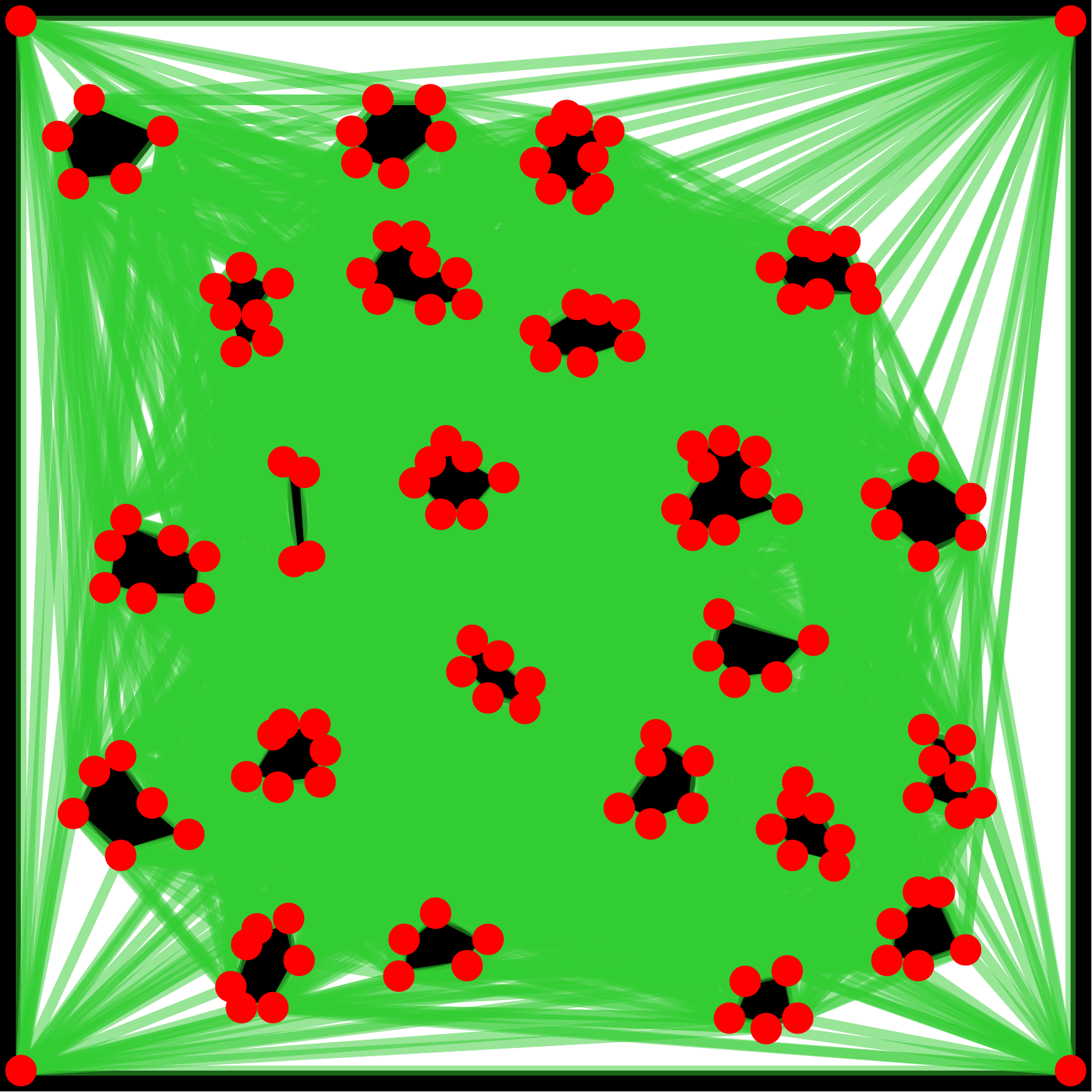}}
        \hfill
        \subfloat{\includegraphics[width=0.328\columnwidth]{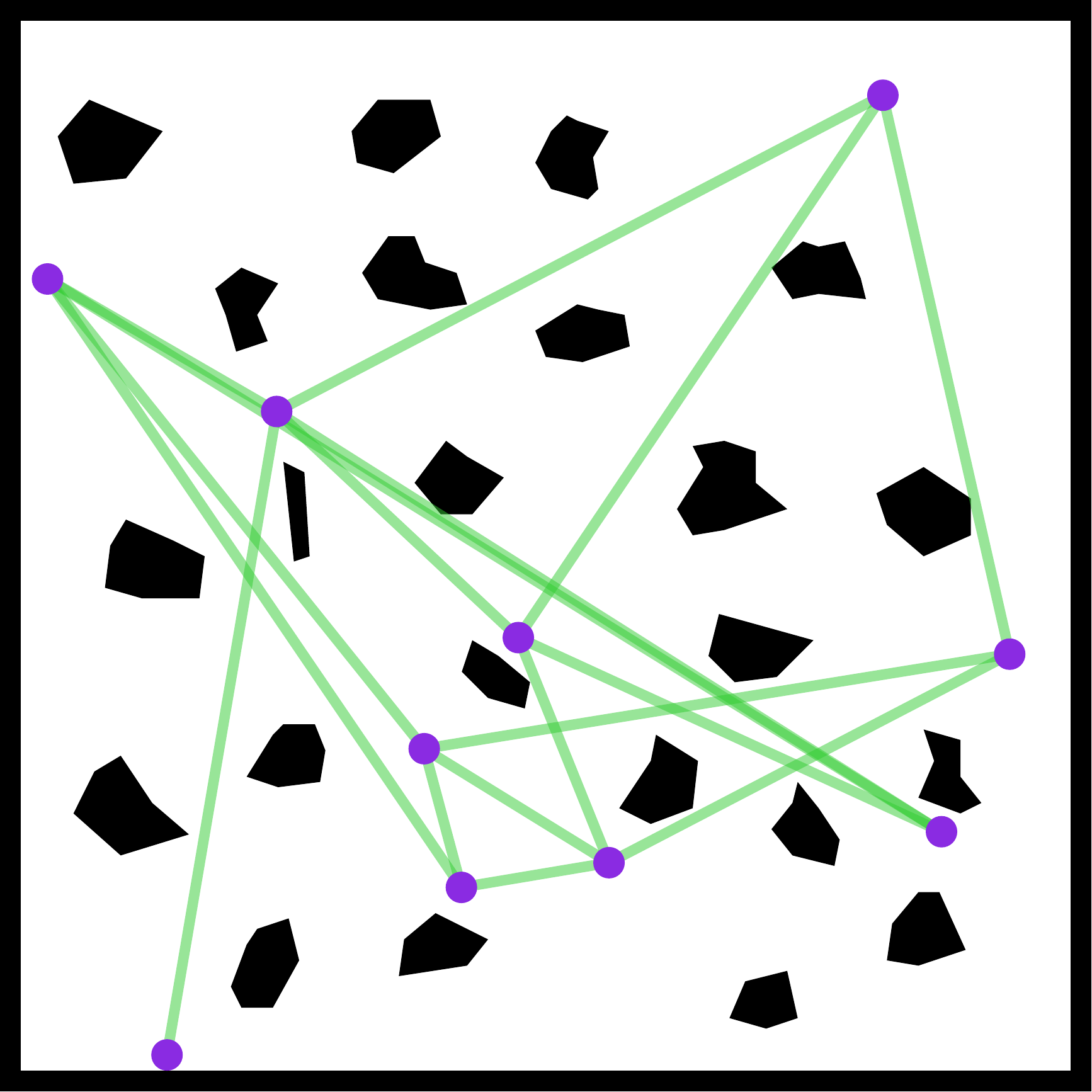}}
        \hfill
        \subfloat{\includegraphics[width=0.328\columnwidth]{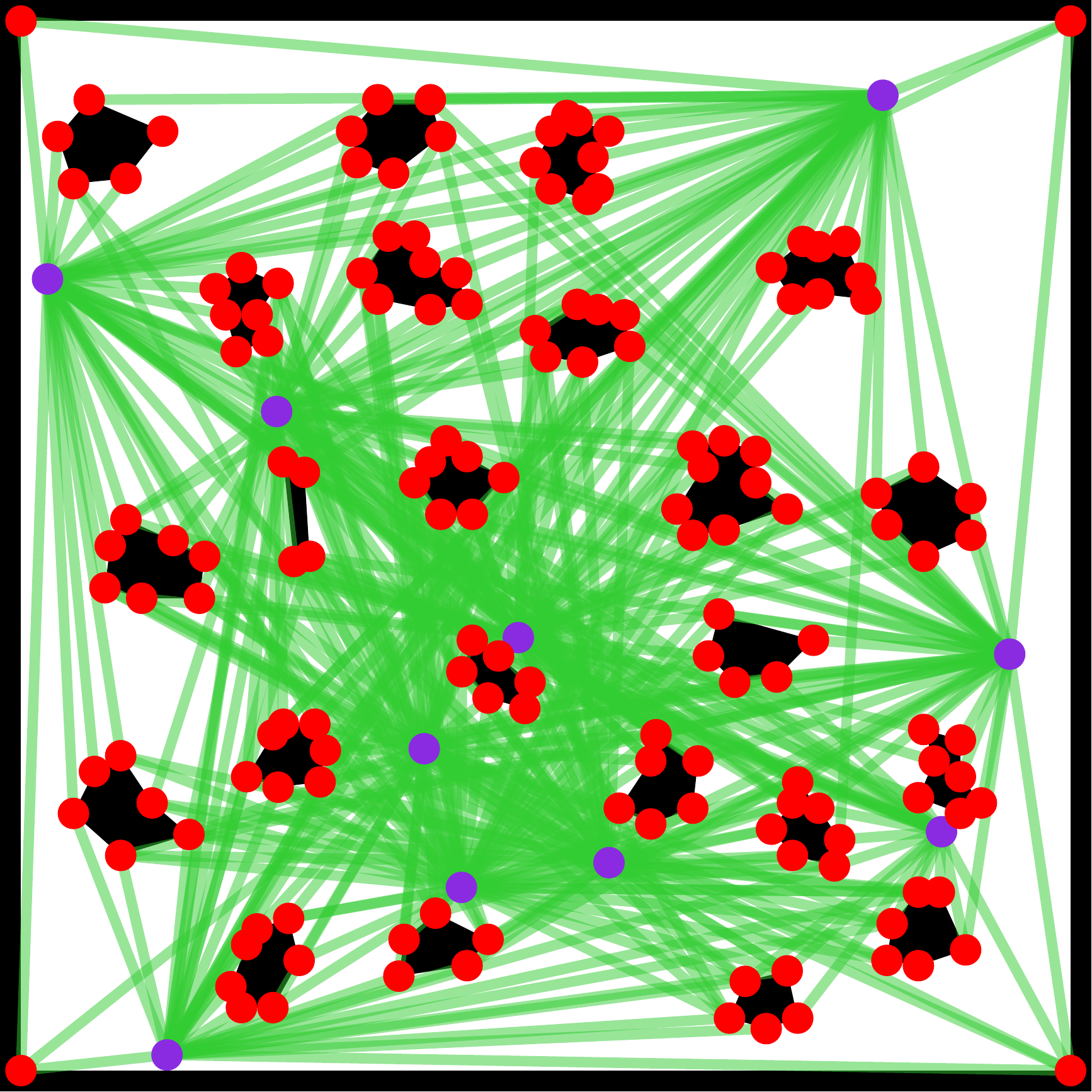}}
        \caption{
            Different visibility queries in the same polygonal environment.
            From left to right, first row: two-points visibility, ray-shooting, visibility polygon.
            Second row: $d$-visibility region, visible vertices, $d$-visible vertices.
            Third row: vertex-vertex visibility graph, point-point visibility graph, vertex-point visibility graph.
        }
        \label{fig:visibility_queries}
    \end{figure}

    This section defines the fundamental concepts of visibility and related terms, illustrated with a visual aid in Fig.~\ref{fig:visibility_queries}.

    \paragraph*{Polygonal Environment}

    A polygonal environment $\mathcal{W}$ is a 2D region bounded by a single outer boundary and zero or more inner boundaries called holes.
    More formally, $\mathcal{W}$ is a non-empty, connected, closed, bounded subset of 2D Euclidean space.
    The notion of obstacles is implicitly included in the definition of $\mathcal{W}$, as $\mathcal{W}$ represents the entire see-through space, while its complement, $\mathbb{R}^2 {\,\setminus\,} \mathcal{W}$, represents the opaque space.
    Furthermore, all boundaries of $\mathcal{W}$ are simple polygons: closed, connected series of pairwise non-intersecting line segments.
    We denote the set of all these segments, united from all the boundaries, as the edges $E_{\mathcal{W}}$ of $\mathcal{W}$, with endpoints referred to as the vertices $V_{\mathcal{W}}$ of $\mathcal{W}$.

    \paragraph*{Convex Polygonal\,/\,Triangular Mesh}

    Any polygonal environment $\mathcal{W}$ can be represented non-uniquely as a convex polygonal mesh $\mathcal{C}$, consisting of convex polygons such that their union forms $\mathcal{W}$, and the intersection of any two polygons is either empty, a single point, or a single line segment.
    We denote the set of all edges formed by the polygons as the edges $E_{\mathcal{C}}$ of $\mathcal{C}$, with endpoints referred to as the vertices $V_{\mathcal{C}}$ of $\mathcal{C}$.
    Unless specified otherwise, we assume that no additional vertices have been added to $\mathcal{C}$, i.e., $V_{\mathcal{C}} {\,=\,} V_{\mathcal{W}}$.
    Moreover, we consider a structure called a convex polygonal graph, which encapsulates $\mathcal{C}$ along with the topological relationships such as adjacency between the polygons, edges, and vertices.
    For simplicity, the convex polygonal graph is denoted by the same symbol $\mathcal{C}$ as its corresponding convex polygonal mesh.
    The process of constructing $\mathcal{C}$ from $\mathcal{W}$ is called convex partitioning.
    A special instance of the convex polygonal mesh is one in which all polygons are triangles.
    Both the triangular mesh and its corresponding triangular graph are denoted as $\mathcal{T}$.
    The process of constructing $\mathcal{T}$ from $\mathcal{W}$ is called triangulation.

    \paragraph*{Point Location Query}

    This query is not necessarily related to visibility but is essential for the operation of T\v{r}iVis.
    It involves determining the location of a point $q {\,\in\,} \mathbb{R}^2$ with respect to $\mathcal{W}$, $\mathcal{C}$, or $\mathcal{T}$.
    The usual objective is to determine whether $q$ lies inside or outside $\mathcal{W}$, and if it lies inside, whether it coincides with one of the vertices $v {\,\in\,} V_{\mathcal{W}}$ or lies on a specific edge $e {\,\in\,} E_{\mathcal{W}}$.
    For $\mathcal{C}$ and $\mathcal{T}$, the objective is similar, but an additional goal is to identify the polygon or triangle, respectively, that contains $q$.

    \paragraph*{Two-Points Visibility Query}

    Two points $q, p {\,\in\,} \mathcal{W}$ are said to be visible to each other if the line segment $\overline{qp}$ lies entirely within $\mathcal{W}$, i.e., $\overline{qp} {\,\subset\,} \mathcal{W}$.
    The query definition is as follows: Given $q, p {\,\in\,} \mathcal{W}$, determine whether $\overline{qp} {\,\subset\,} \mathcal{W}$.

    \paragraph*{Ray Shooting}

    This visibility-related query aims to find the first intersection point of a half-line, called a ray, with the boundary of $\mathcal{W}$.
    Given a point $q {\,\in\,} \mathcal{W}$ and a directional vector $u {\,\in\,} \mathbb{R}^n$, which defines the ray, the objective is to find the point $p {\,=\,} \argmin_{\lambda {\,\geq\,} 0} \{ q {\,+\,} \lambda u {\,\mid\,} q {\,+\,} \lambda u {\,\in\,} \partial \mathcal{W} \}$, where $\partial \mathcal{W}$ is the boundary of $\mathcal{W}$.
    Note that $p$ is always defined because $\mathcal{W}$ is a closed, bounded region, while the ray starts inside $\mathcal{W}$ and extends indefinitely.
    Additionally, the above expression implies that $q$ and $p$ are visible to each other, i.e., $\overline{qp} {\,\subset\,} \mathcal{W}$.

    \paragraph*{Visibility Region\,/\,Polygon}

    The visibility region is the set of all points visible from a given point $q$, defined as $\mathcal{V}(q) {\,=\,} \{ p {\,\in\,} \mathcal{W} {\,\mid\,} \overline{qp} {\,\subset\,} \mathcal{W} \}$.
    In polygonal environments, visibility regions take the form of polygons.
    These polygons are usually simple, except in cases where one-dimensional structures, known as antennas, are created.
    Antennas occur when the query point aligns with two vertices of the environment, restricting visibility from opposite sides.

    \paragraph*{Visible Points\,/\,Vertices Query}

    The visible points query aims to identify a subset of a given finite set of points $P {\,\subset\,} \mathcal{W}$ that are visible from a given query point $q {\,\in\,} \mathcal{W}$.
    Its definition resembles that of the visibility region, but the output domain is confined to $P$: $\mathcal{V}_P(q) {\,=\,} \{ p {\,\in\,} P {\,\mid\,} \overline{qp} {\,\subset\,} \mathcal{W} \}$.
    In polygonal environments, we can consider a special case of the visible points query, called the visible vertices query, where $P {\,\subset\,} V_{\mathcal{W}}$.
    We denote the output in this case as $\mathcal{V}_V(q)$.

    \paragraph*{Visibility Graph\,/\,Subgraphs}

    The visibility graph is a structure used to represent the pairwise visibility relationships among a given finite set of points $Q {\,\subset\,} \mathcal{W}$.
    It is defined as a simple undirected graph $G_Q {\,=\,} (V_Q, E_Q)$, where $V_Q {\,=\,} Q$ and $E_Q {\,=\,} \{ \{ q, p \} {\,\mid\,} q, p {\,\in\,} Q {\,\land\,} q {\,\neq\,} p {\,\land\,} \overline{qp} {\,\subset\,} \mathcal{W} \}$ are the vertex and edge sets, respectively.
    In polygonal environments, we can classify three types of visibility subgraphs by writing the set of visibility graph query points as $Q {\,=\,} V {\,\cup\,} P$, where $V {\,\subset\,} V_{\mathcal{W}}$ is a set of vertices, $P {\,\subset\,} V_{\mathcal{W}}$ is a set of non-vertex points, and $V {\,\cap\,} P {\,=\,} \emptyset$.
    Then, the vertex-vertex visibility graph, denoted as $G_V {\,=\,} (V_V, E_V)$, is the subgraph of $G_Q$ induced by $V$.
    The point-point visibility graph, denoted as $G_P {\,=\,} (V_P, E_P)$, is the subgraph of $G_Q$ induced by $P$.
    Lastly, the vertex-point visibility graph is defined as $G_{V\!P} {\,=\,} (V_{V\!P}, E_{V\!P})$, where $V_{V\!P} {\,=\,} V_Q$ and $E_{V\!P} {\,=\,} E_Q {\,\setminus\,} E_V {\,\setminus\,} E_P$.
    Note the following relationships: $V_Q {\,=\,} V_{V\!P} {\,=\,} V_V {\,\cup\,} V_P$, $V_V {\,\cap\,} V_P {\,=\,} \emptyset$, $E_Q {\,=\,} E_V {\,\cup\,} E_P {\,\cup\,} E_{V\!P}$, and $E_V {\,\cap\,} E_P {\,=\,} E_V {\,\cap\,} E_{V\!P} {\,=\,} E_P {\,\cap\,} E_{V\!P} {\,=\,} \emptyset$.

    \paragraph*{Limited Visibility}

    Visibility may be subject to additional constraints.
    In particular, we can consider a limited visibility range $d$: two points $q$ and $p$ in $\mathcal{W}$ are considered $d$-visible to each other if the line segment $\overline{qp}$ lies entirely within $\mathcal{W}$ and has a length no greater than $d$, i.e., $\overline{qp} {\,\subset\,} \mathcal{W} {\,\land\,} \|\overline{qp}\| {\,\leq\,} d$.
    The set of all points $d$-visible from a given point $q$ is called the $d$-visibility region and is defined as $\mathcal{V}_d(q) {\,=\,} \{ p {\,\in\,} \mathcal{W} {\,\mid\,} \overline{qp} {\,\subset\,} \mathcal{W} {\,\land\,} \|\overline{qp}\| {\,\leq\,} d \}$.
    Note that the $d$-visibility region may no longer be a polygon, as it may contain circular arcs due to the limited visibility range.
    Similarly, we could define the $d$-visibility graph, $d$-visible points query, or $d$-visibility ray shooting (with a Null result if $\|\overline{qp}\| > d$).

    \section{Visibility Algorithms}
    \label{sec:algorithms}

    This paper does not introduce new algorithms, nor does it aim to offer an exhaustive review of existing visibility algorithms for all queries defined in Sec.~\ref{sec:definitions}.
    Instead, the presented library is centered around the triangular expansion algorithm (TEA), which has been adapted to handle all these visibility queries in polygonal environments.
    To provide a partial background, we present a literature review of algorithms for computing visibility polygons in Sec.~\ref{subsec:background}.
    Following this, the TEA is elaborated upon in Sec.~\ref{subsec:tea}, and the section concludes with a description of the T\v{r}iVis-specific adaptations for handling other queries in Sec.~\ref{subsec:tea_adaptations}.

    \subsection{Literature Background on Visibility Polygons}
    \label{subsec:background}

    Computing visibility polygons is a well-explored area within computational geometry originating in the late 1970s~\cite{Davis1979}.
    Since then, it has received continuous attention from numerous researchers, primarily focusing on the theoretically proven complexity of the algorithms rather than their practical implementation and experimental validation.
    Initially, the emphasis was on simple polygons, aiming to reduce the query time to $\mathcal{O}(n)$, where $n$ denotes the number of vertices in the polygon.
    The first correct $\mathcal{O}(n)$ solution was proposed by Joe and Simpson~\cite{Joe1987}.

    Subsequently, attention shifted towards polygons with holes until Heffernan and Mitchell~\cite{Heffernan1995} introduced the optimal $\mathcal{O}(n {\,+\,} h\log h)$ algorithm, where $h$ signifies the number of holes, surpassing the previously leading $\mathcal{O}(n\log n)$-time rotational sweep algorithm (RSA) by Asano~\cite{Asano1985}.
    In subsequent decades, numerous algorithms emerged for polygons with holes, offering diverse trade-offs between preprocessing time and query time.
    For example, Zarei and Ghodsi~\cite{Zarei2008} presented an algorithm with $\mathcal{O}(n^3\log n)$ preprocessing time and $\mathcal{O}(K {\,+\,} \min(h,K)\log n)$ query time; Inkulu and Kapoor~\cite{Inkulu2009} introduced a method with $\mathcal{O}(n^2\log n)$ preprocessing time and $\mathcal{O}(K\log^{2}n)$ query time; and Chen and Wang~\cite{Chen2015} devised an approach with $\mathcal{O}(n^2\log n)$ preprocessing time and $\mathcal{O}(K {\,+\,} \log^{2}n{\,+\,}h\log(n/h))$ query time.

    As previously mentioned, the research discussed thus far has primarily focused on theoretical aspects, with limited attention given to practical implementation, creating a notable gap in the field.
    This gap was addressed by Bungiu et~al.~\cite{Bungiu2014}, who implemented Joe and Simpson's algorithm for simple polygons~\cite{Joe1987} and Asano's RSA for polygons with holes~\cite{Asano1985} within the then-new CGAL 2D Visibility package.

    Since the remaining algorithms presented in the literature were deemed too complex for practical implementation, Bungiu et~al.~\cite{Bungiu2014} also introduced their own solution, the TEA, which they integrated into the same package.
    Although the TEA, with a worst-case query time of $\mathcal{O}(nh)$, does not rank among the theoretically fastest algorithms, the authors demonstrated that it performs two orders of magnitude faster than the RSA in real-world scenarios, establishing it as the favored algorithm for computing visibility polygons in practice.
    Although Bungiu et~al.'s work is significant, we faced several difficulties while using the CGAL 2D Visibility package for our research, which motivated the development of T\v{r}iVis.
    These difficulties are apparent from our evaluation of the package in Sec.~\ref{sec:performance}.

    It is worth noting that Xu and G{ü}ting~\cite{Xu2015} independently\footnote{
        Xu and G{\"u}ting's manuscript was submitted for publication nearly six months before Bungiu et al.'s manuscript appeared on ArXiv, which, to our knowledge, remains unpublished.
        Therefore, it is highly likely that these two independent research groups were unaware of each other's work.
    } developed an equivalent algorithm to the TEA for the visible vertices query.
    However, their work has received little to no attention compared to Bungiu et~al.'s~\cite{Bungiu2014}, as evidenced by the Google Scholar citation count.

    \subsection{Triangular Expansion Algorithm}
    \label{subsec:tea}

    We base our description on~\cite{Bungiu2014}, as computing visibility polygons is more relevant to our evaluation in Sec.~\ref{sec:performance}.
    The following description assumes that the input polygonal environment $\mathcal{W}$ has been triangulated into $\mathcal{T}$ and that the query point $q {\,\in\,} \mathcal{W}$ has been located inside a triangle $\Delta {\,\in\,} \mathcal{T}$.

    \paragraph*{Basic Idea and Features}

    The underlying idea is simple: The TEA computes the visibility polygon $\mathcal{V}(q)$ by recursively traversing neighboring triangles, starting from $\Delta$.
    The first notable feature is that it exclusively traverses triangles visible from the query point, making it output-sensitive to some extent.
    However, it may also traverse triangles that do not contribute directly to the boundary of the resulting visibility polygon~\cite{Bungiu2014}, rendering it only partially output-sensitive.
    The second significant feature is the simplicity of the operations during mesh traversal.
    These operations essentially involve answering two orientation predicates per triangle traversal and computing at most two ray-segment intersections when encountering an edge of $\mathcal{W}$.

    \paragraph*{Detailed Operation of the TEA}

    Starting at $\Delta$, a recursive expansion procedure is initiated for each of its edges, gradually forming a set of restricted views around $q$.%
    \footnote{
        An animated visualization of the TEA operation is available at \url{https://www.youtube.com/watch?v=gKSA6lxVTKw}.
    }
    The recursive procedure expands along the current edge to the neighboring triangle, restricting the view from $q$ between the edge's endpoints.
    In the new triangle, the other two edges are considered as candidates for the next recursion call only if they intersect the current restricted view from $q$.
    The new recursion call is eventually initiated for any of those candidate edges that neighbor another triangle.
    Otherwise, the edge is identified as a boundary edge of $\mathcal{W}$, and the current view from $q$ is ultimately restricted between its endpoints, stored, and the recursion does not propagate further from that point.
    Once no further expansions are possible, the resulting $\mathcal{V}(q)$ is the union of all the restricted views formed around $q$.
    Additionally, an efficient TEA implementation takes advantage of pre-ordering the neighbor information in either clockwise (cw) or counterclockwise (ccw) order.
    The expansions then rotate around $q$ while adhering to the same order, naturally forming the union of the restricted views.

    \paragraph*{Vertex Query}

    When the query point $q$ coincides with one of the mesh vertices, a special case arises where the TEA must consider the union of all triangles incident to that vertex.
    This union forms the basis of the resulting visibility polygon, similarly to the initial triangle $\Delta$ in the general case.
    The expansion procedure is initiated for each outer edge of that union, where an outer edge is defined as an edge that is not shared with another triangle in the union.
    Otherwise, the operation is identical to the one described above.

    \paragraph*{Complexity Analysis}

    As stated by Bungiu et~al.~\cite{Bungiu2014}, the worst-case query time is $\mathcal{O}(n^2)$, where $n {\,=\,} |V_{\mathcal{T}}| {\,=\,} |V_{\mathcal{W}}|$.
    This is because the recursion may split into two views $\mathcal{O}(n)$ times, and each view may reach $\mathcal{O}(n)$ triangles.
    However, splits into two views that independently reach the same triangle may only occur at the holes of $\mathcal{W}$.
    This implies that the worst-case query time is rather $\mathcal{O}(nh)$.%
    \footnote{
        The possibility for two independent views to reach the same triangle by splitting at a hole is an essential property of the TEA, which seems to not have been realized by the other independent authors of the TEA, Xu and G{ü}ting \cite{Xu2015}.
        Therefore, their complexity analysis, which states the worst-case query time to be $\mathcal{O}(n)$, is incorrect.
    }

    \paragraph*{Preprocessing}

    When the input for the TEA is not the triangular mesh $\mathcal{T}$, but instead the polygonal environment $\mathcal{W}$ or the convex polygonal mesh $\mathcal{C}$, the TEA query phase must be preceded by the respective preprocessing phase.
    The preprocessing involves the triangulation of $\mathcal{W}$ or individual polygons in $\mathcal{C}$.
    For example, this can be achieved by the constrained Delaunay triangulation (CDT), for which exist algorithms running in $\mathcal{O}(n \log n)$~\cite{Chew1987} time.

    \paragraph*{Optimal Mesh}

    TEA authors have not examined how the choice of triangular mesh affects the query performance of the TEA\@.
    To fill this gap, we address this issue in our prior research~\cite{Mikula2024}, where we derive the optimal triangular mesh under the assumption that query points are uniformly distributed over $\mathcal{W}$.
    Unfortunately, constructing this optimal mesh involves solving an NP-hard problem.
    Thus, we propose a parametric heuristic approach that balances mesh quality and preprocessing time.
    We assess the TEA's performance with the approximate optimal mesh, denoted as MinVT, on a dataset of complex polygonal environments (the same dataset as used in this paper) and compare it with the performance of the TEA with CDT\@.
    Our experiments show that the MinVT mesh can improve the query time by 12--16\% compared to CDT, at the expense of 9--212 seconds of preprocessing, depending on the chosen parametrization.
    While this improvement may not meet our initial expectations, it is still significant enough to justify the use of the MinVT mesh in applications where preprocessing time is not a critical factor.
    However, for simplicity, in this paper, we opt for the CDT mesh.

    \subsection{Adaptations for the Other Visibility Queries}
    \label{subsec:tea_adaptations}

    In T\v{r}iVis, the TEA is the primary algorithm for computing visibility in polygonal environments.
    Initially, the library was developed to compute visibility polygons but has since been adapted to handle other visibility queries from Sec.~\ref{sec:definitions}.
    A brief overview of these adaptations follows.

    Adapting the TEA for the visible vertices query involves a straightforward modification, where the output is now a collection of the traversed triangles' vertices $\mathcal{V}_V(q)$ visible from the query point $q$.
    Furthermore, there is no need to compute any intersections with the boundary of $\mathcal{W}$, which reduces the computational effort.
    The visible points query follows a similar process, but it requires precomputed point location queries with respect to $\mathcal{T}$ for the input set of points $P$.
    These input points are then stored within the corresponding triangle containing them.
    During the query phase, the algorithm looks up the stored points for each traversed triangle and assesses whether they fall within the current restricted view from the query point.
    The subset of $P$ containing points that have been visible at some point during the traversal is returned as the output at the end of the expansion procedure.

    In the context of the two-points visibility query, we denote one of the query points as the source~$q$ and the other as the target~$p$.
    Similar to other queries, T\v{r}iVis initiates the expansion from $q$.
    However, unlike the previously mentioned queries, it does not need to expand in all directions; rather, it expands directly towards $p$.
    This means that only triangles along the line segment $\overline{qp}$ are traversed, which implies that no views are split at holes and the worst-case query time is $\mathcal{O}(n)$ for this query.
    If the target is reached before encountering the boundary of $\mathcal{W}$, indicating that the target is visible from the source, the algorithm returns true; otherwise, it returns false.
    By adjusting this approach, we can derive the ray-shooting query, where no target is considered, and only the direction vector $u$ guides the expansions.
    In this case, the expansion procedure terminates as soon as the first intersection with the boundary of $\mathcal{W}$ is found, and the intersection point is returned.

    Computing the visibility graph for a set of query points $Q = V {\,\cup\,} P$ is a more complex task, as it involves computing the three types of visibility subgraphs and, if desired, merging them into a single graph.
    For the vertex-vertex and vertex-point visibility graph, the visible vertices query is executed for each vertex in $V$ and each point in $P$, respectively.
    For the point-point visibility graph, the visible points query is executed for each point in $P$.

    T\v{r}iVis can also handle $d$-visibility queries by incorporating $d$ as an additional input to the TEA, resulting in a variant we denote as d-TEA\@.
    A branch of the d-TEA's expansion procedure terminates prematurely if the distance from the query point to the edge being expanded exceeds $d$.
    This can significantly reduce the number of traversed triangles, especially for small values of $d$ relative to the expected distance between two points in $\mathcal{W}$ that are visible to each other.
    To ensure correctness, the query outputs are also adjusted to account for the distance constraint, typically involving straightforward distance checks between the query point and the output points.
    The most complex case arises with the $d$-visibility region, where the output of the expansion procedure, $\mathcal{V}'_d$, is a superset of the correct result, $\mathcal{V}_d \subset \mathcal{V}'_d$, and the final output is obtained by intersecting $\mathcal{V}'_d$ with a circle centered at the query point with radius $d$.

    \section{Library Design and Usage}
    \label{sec:design}

    This section outlines T\v{r}iVis's design, covering robustness strategies (Sec.~\ref{subsec:robustness}), the visibility query scheme (Sec.~\ref{subsec:structure}), and point location implementation (Sec.~\ref{subsec:point_location}).
    We also review the library's internal dependencies in Sec.~\ref{subsec:dependencies} and provide a usage code snippet in Sec.~\ref{subsec:usage}.

    \subsection{Robustness Strategies}
    \label{subsec:robustness}

    T\v{r}iVis relies on floating-point arithmetic and incorporates a unique combination of adaptive robust geometry predicates~\cite{RichardShewchuk1997} and $\epsilon$-geometry~\cite{Fortune1993}.
    The predicates are employed in T\v{r}iVis's expansion procedure, ensuring consistent answers to orientation queries.
    This prevents the algorithm from encountering infinite loops and considerably enhances its performance compared to using the non-adaptive versions~\cite{RichardShewchuk1997}.

    Additionally, $\epsilon$-geometry is employed to address persistent issues despite the utilization of robust predicates.
    In particular, we have encountered the following two issues during the development of T\v{r}iVis.
    First, due to numerical inaccuracies, computing the correct intersection points for visibility queries when the query point is near some vertex of $\mathcal{W}$ has been difficult.
    To address this, we implemented the so-called vertex snapping technique controlled by the $\epsilon_2$ value, as described in Sec.~\ref{subsec:structure}.

    Second, we faced instances where the robust predicates failed to determine that a point lies on a triangle edge, despite the point being computed as the midpoint of that edge.
    For example, considering triangle vertices represented as vectors $a$ and $b$, with $c$ computed as $c {\,=\,} (a {\,+\,} b){\,/\,}2$ in floating-point representation, the robust geometric predicates indicate that $c$ lies outside the triangle in high percentage of cases.
    While this behavior is not necessarily a problem of the robust predicates themselves, it is likely undesirable from the user's perspective.
    To mitigate this issue, we implemented the point location query as a combination of exact and inexact arithmetic, as described in Sec.~\ref{subsec:structure}, where the inexact arithmetic is controlled by a sequence of increasing $\epsilon_1$ values.

    \subsection{General Scheme for Answering Visibility Queries}
    \label{subsec:structure}

    The T\v{r}iVis library consists of a set of C++ classes and functions, with its primary class being \textsc{Trivis}.
    This class serves as the core of the library, responsible for executing visibility queries, managing query input and output, and preprocessing the polygonal environment.
    Query execution relies on the TEA and its adaptations, detailed in Sec.~\ref{subsec:tea} and Sec.~\ref{subsec:tea_adaptations}, respectively, and follows a generic scheme outlined in Alg.~\ref{alg:generic_query}.

    \begin{algorithm}
        \caption{Generic Visibility Query Scheme}
        \label{alg:generic_query}
        \begin{algorithmic}[1]%
            \Function{X-Query}{$q$, $\mathcal{E}_1$, $\epsilon_2$, $\dots$}
                \label{alg:generic_query:l1}%
                \State $\Delta \gets {}$\Call{FindTriangle}{$q$}\label{alg:generic_query:l2}%
                \While{$\Delta \textbf{ is }\text{Null}$}
                    \label{alg:generic_query:l3}%
                    \If{$\mathcal{E}_1 \textbf{ is }\text{Empty}$}
                        \label{alg:generic_query:l4}%
                        \State \Return $\text{Null}$\label{alg:generic_query:l5}%
                    \EndIf%
                    \State $\epsilon_1 \gets \Call{PopSmallest}{\mathcal{E}_1}$\label{alg:generic_query:l6}%
                    \State $\Delta \gets {}$\Call{FindTriangleWithEpsilon}{$q$, $\epsilon_1$}\label{alg:generic_query:l7}%
                \EndWhile
                \For{$v \in V_{\!\Delta}$}
                    \label{alg:generic_query:l8}%
                    \If{$\|\overline{vq}\| {\,\leq\,} \epsilon_2$}
                        \label{alg:generic_query:l9}%
                        \State \Return \Call{X-ExpandVertex}{$v$, $\dots$}\label{alg:generic_query:l10}%
                    \EndIf%
                \EndFor%
                \State \Return \Call{X-ExpandTriangle}{$q$, $\Delta$, $\dots$}\label{alg:generic_query:l11}%
            \EndFunction
        \end{algorithmic}
    \end{algorithm}

    Each visibility query requires the query point $q$ as mandatory input, except for visibility graphs, which are constructed through multiple calls to this type of query function.
    Additionally, the query function accepts specific arguments depending on the query type, such as the target point $p$ for the two-points visibility query, directional vector $u$ for the ray-shooting query, and the set of points $P$ for the visible points query, or the distance $d$ for all the $d$-visibility queries.
    The query function also accepts a set of $\epsilon$ values (a sequence $\mathcal{E}_1$ and a single $\epsilon_2$), which enhance the robustness and control the behavior of the query.
    The motivation behind these $\epsilon$ values is discussed in Sec.~\ref{subsec:robustness}; here, we focus on the precise logic of their usage.

    The generic query function in Alg.~\ref{alg:generic_query} first attempts to find the triangle $\Delta$ containing $q$ using the robust geometric predicates (line~\ref{alg:generic_query:l2}).
    If this fails, it resorts to $\epsilon$-geometry, employing the sequence of increasing $\epsilon_1$ values to locate $\Delta$ (lines~\ref{alg:generic_query:l3}--\ref{alg:generic_query:l7}).
    If $\Delta$ is still not found, it is interpreted that $q$ lies outside $\mathcal{W}$, and the function returns Null (line~\ref{alg:generic_query:l5}) to indicate this.
    If $\Delta$ is found, the function checks whether $q$ is near any vertices of $\Delta$ using the $\epsilon_2$ value (lines~\ref{alg:generic_query:l8}--\ref{alg:generic_query:l9}).
    If so, the vertex query is executed (line~\ref{alg:generic_query:l10}), expanding all outer edges of the union of triangles incident to the vertex.
    When $\epsilon_2 > 0$ and $q {\,\neq\,} v$, the operation at lines~\ref{alg:generic_query:l8}--\ref{alg:generic_query:l10} can be interpreted as vertex snapping, where $q$ is snapped to $v$.
    Otherwise, the standard expansion procedure is initiated, expanding all edges of $\Delta$.
    The procedure is recursive, and the function returns the query output once it eventually terminates (line~\ref{alg:generic_query:l10} or~\ref{alg:generic_query:l11}).

    \subsection{Bucketing for Point Location}
    \label{subsec:point_location}

    The point location query identifies the triangle $\Delta$ containing the query point, which is the first step of Alg.~\ref{alg:generic_query}.
    In T\v{r}iVis, we implement the bucketing technique from~\cite{Edahiro1984} to achieve an average time complexity of $\mathcal{O}(1)$.
    This technique preprocesses the triangular mesh $\mathcal{T}$ by subdividing the bounding box of $\mathcal{W}$ into square cells called buckets, each storing the triangles intersecting it.
    During the query phase, the bucket containing the query point is found by rounding the point's coordinates.
    Finally, the bucket's triangles undergo the point-in-triangle test, resulting in the identification of triangle $\Delta$.
    The test is performed exactly using the robust predicates at line~\ref{alg:generic_query:l2} of Alg.~\ref{alg:generic_query}, and using the $\epsilon_1$ values at line~\ref{alg:generic_query:l7}.

    \subsection{Internal Dependencies}
    \label{subsec:dependencies}

    T\v{r}iVis, implemented in C++17, is self-contained, ensuring that its core functionality is independent of external libraries.
    However, internally, it depends on some third-party software that is freely available for private, research, and institutional use, and is integrated into the library's source code.
    Most importantly, T\v{r}iVis relies on Robust-Predicate\footnote{Available at \url{https://github.com/dengwirda/robust-predicate}.} for the adaptive robust geometry predicates, and Triangle\footnote{Available at \url{https://www.cs.cmu.edu/~quake/triangle.html}.}~\cite{Shewchuk1996, Shewchuk2002} for computing triangular meshes.
    Furthermore, Clipper2\footnote{Available at \url{https://github.com/AngusJohnson/Clipper2}.} is integrated for polygon-related operations such as clipping and offsetting.
    Although the core functionality of T\v{r}iVis does not necessarily require Clipper2, it is utilized in the library's evaluation in Sec.~\ref{sec:performance}.
    The remaining core functionality of T\v{r}iVis, as outlined in this paper, represents original contributions from the authors.

    \subsection{Library Usage}
    \label{subsec:usage}

    T\v{r}iVis offers a user-friendly interface through the \textsc{Trivis} class for executing all the visibility queries defined in Sec.~\ref{sec:definitions}.
    Fig.~\ref{fig:code_snippet} shows a code snippet for computing a $d$-visibility region from a query point $q$.
    This snippet is simplified for brevity and requires the user to provide the environment data, the query point, and set the visibility range $d$ (the latter two have default values).
    The output is a polygonal approximation of the $d$-visibility region, with circular arcs sampled using line segments at an angle no greater than $\pi/180$.
    For detailed usage and complete example projects, including visualization, refer to the library's \texttt{README.md} on the GitHub repository at \url{https://github.com/janmikulacz/trivis}.

    \begin{figure}
        \begin{center}%
            \begin{minipage}{0.95\columnwidth}%
                \begin{lstlisting}[caption=]
#include "trivis/trivis.h"
int main() {
  using namespace trivis;
  geom::PolyMap environment; // TODO: Fill the environment.
  Trivis vis{std::move(environment)}; // Construct the visibility object.
  geom::FPoint q{0.0, 0.0}; // TODO: Fill the query point q.
  std::optional<double> d = 10.0; // TODO: Set the visibility range d.
  // Locate the query point q:
  std::optional<Trivis::PointLocationResult> pl = vis.LocatePoint(q);
  if (!pl.has_value()) return EXIT_FAILURE; // q is outside the environment.
  // Compute the d-visibility region from q:
  AbstractVisibilityRegion abs = vis.VisibilityRegion(q, pl.value(), d);
  // Note: abs is an intermediate representation w/o computed intersections.
  RadialVisibilityRegion reg = vis.ToRadialVisibilityRegion(abs);
  if (d.has_value()) reg.IntersectWithCircleCenteredAtSeed(d);
  // Optional post-processing:
  if (d.has_value()) reg.SampleArcEdges(M_PI / 180.0);
  geom::FPolygon poly_approx = reg.ToPolygon();
  // TODO: Do something with the polygonal approx. of the visibility region.
  return EXIT_SUCCESS;
}               \end{lstlisting}%
            \end{minipage}%
        \end{center}%
        \vspace{-1em}%
        \caption{Code snippet for computing a $d$-visibility region from a query point $q$ using T\v{r}iVis.}%
        \label{fig:code_snippet}%
    \end{figure}

    \section{Performance Evaluation}
    \label{sec:performance}

    T\v{r}iVis's performance evaluation focuses on computing visibility polygons in polygonal environments and comparing it with other implementations in terms of runtime behavior and computational time.
    The evaluation methodology is detailed in Sec.~\ref{subsec:methodology}, and the results are presented in Sec.~\ref{subsec:results}.

    \subsection{Methodology}
    \label{subsec:methodology}

    \paragraph*{Evaluated Implementations}

    Among the other evaluated implementations are the RSA~\cite{Asano1985} and TEA~\cite{Bungiu2014} from the CGAL 2D Visibility package, version 5.6.%
    \footnote{
        Available at \url{https://www.cgal.org/}.
    }
    These are evaluated with exact predicates and either exact (CE) or inexact (CI) construction kernels.
    In addition to CGAL, VisiLibity1\footnote{
        Available at \url{https://karlobermeyer.github.io/VisiLibity1/}.
    }
    is included, with its documentation claiming an average $\mathcal{O}(n\log n)$ query performance with no preprocessing; however, the specific algorithm employed in VisiLibity1 is not disclosed in the documentation.

    \paragraph*{Map Dataset}

\begin{figure}
        \centering
        \subfloat{\includegraphics[height=0.35\columnwidth]{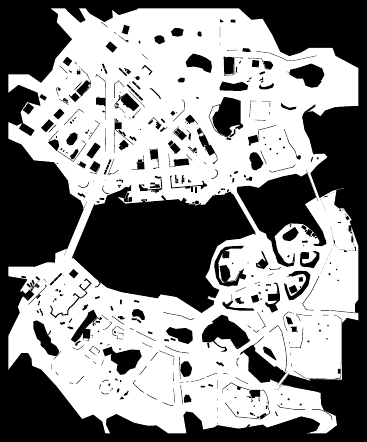}}
        \hfill
        \subfloat{\includegraphics[height=0.35\columnwidth]{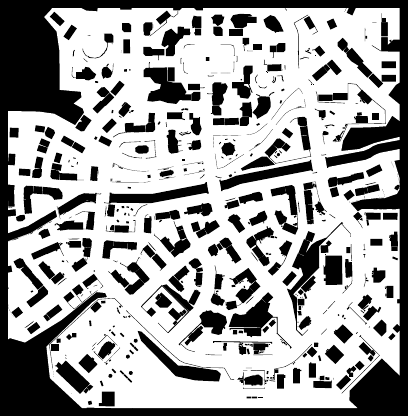}}
        \hfill
        \subfloat{\includegraphics[height=0.35\columnwidth]{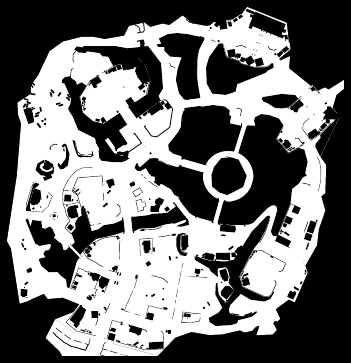}}
        \caption{Examples of polygonal environments from the Iron Harvest dataset.}
        \label{fig:iron_harvest}
    \end{figure}

    Our benchmark environments are derived from a dataset of 35 maps from the Iron Harvest video game, introduced in~\cite{Harabor2022}, and were chosen specifically for their scale and complexity.
    A notable advantage of this dataset is that it includes three representations of the maps---polygonal, mesh, and grid---while ensuring that the maps maintain the same topology.
    This provides benefits over robotic datasets, which are either limited to grid representations or, if polygonal, are usually too small and lack the complexity of the Iron Harvest maps.
    We selected the polygonal representation for our purposes.
    To ensure each environment is well-formed and connected, we preprocess it by selecting the largest polygon to represent the outer boundary and incorporating all its holes.
    We discard any additional disconnected artifacts, such as polygons that are not connected to the main polygon with holes or those fully enclosed within the holes.
    The resulting polygonal environments are characterized by thousands of vertices and dozens to hundreds of holes, typically spanning across 400$\times$400 units.
    We assessed the validity of the resulting environments using CGAL's \textsc{CGAL::Arrangement\_2::is\_valid} method, excluding a single invalid instance, \textsc{scene\_sp\_cha\_02}, from the dataset.
    To illustrate their complexity, Fig.~\ref{fig:iron_harvest} showcases three instances used in the evaluation.

    \paragraph*{Query Points}

    We generated six sets of 1,000 query points for all maps: \textit{In}, \textit{BB}, \textit{Ver}, \textit{NearV}, \textit{Mid}, and \textit{NearM}.
    The \textit{In} set contains points uniformly distributed inside $\mathcal{W}$, and the \textit{BB} set within its bounding box.
    The \textit{Ver} set includes randomly selected vertices of $\mathcal{W}$, while the \textit{Mid} set consists of midpoints of edges $\{a, b\} \in \mathcal{T}$, where $\mathcal{T}$ is the CDT of $\mathcal{W}$.
    The \textit{NearV} and \textit{NearM} sets are based on \textit{Ver} and \textit{Mid} with added normally distributed noise ($\sigma$ randomly selected from $\{10^{-15}, 10^{-14}, \dots, 10^{-1}\}$).
    For the 34 valid maps, this results in 34,000 points per set, totaling 204,000 query points.

    \paragraph*{Experimental Setup}

    The T\v{r}iVis's epsilon values are set to $\mathcal{E}_1 {\,=\,} (10^{-18},\allowbreak 10^{-17},\allowbreak \dots,\allowbreak 10^{-9})$ and $\epsilon_2 {\,=\,} 10^{-12}$, and the bucket size for the point location query is set to 1 unit.
    In VisiLibity1, the same value of $\epsilon_2$ is applied for edge and vertex snapping, as recommended in the documentation.
    Furthermore, all implementations are single-threaded, compiled in Release mode using GCC 12.3.0, and tested on a personal laptop, the Lenovo Legion 5 Pro 16ITH6H, with an Intel Core i7-11800H (4.60GHz), 16GB of RAM, and running Ubuntu 20.04.6 LTS\@.

    \paragraph*{Runtime Behaviors|Definitions}

    We use CGAL's TEA with exact predicates and constructions (CGAL-TEA-CE) as the reference implementation $\mathcal{R}$.
    For the tested implementation $\mathcal{A}$, we define the following exclusive runtime behaviors:
    \begin{itemize}
        \item \textit{Crash:} $\mathcal{A}$ crashes due to a segmentation fault, unhandled exception or is killed by the OS\@.
        \item \textit{Inf:} $\mathcal{A}$ does not finish within its usual time frame likely due to an infinite loop and must be killed manually.
        \item \textit{NoRef:} $\mathcal{A}$ finishes, but $\mathcal{R}$'s output is unavailable for comparison due to \textit{Crash} or \textit{Inf}.
        \item \textit{Null:} Both $\mathcal{A}$ and $\mathcal{R}$ return Null.
        \item \textit{A0R1:} $\mathcal{A}$ returns Null when $\mathcal{R}$ returns a non-Null.
        \item \textit{A1R0:} $\mathcal{A}$ returns a non-Null when $\mathcal{R}$ returns Null.
        \item \textit{Same:} $\mathcal{A}$ and $\mathcal{R}$ return the same non-Null output.
        This is determined by computing the XOR of the two outputs using Clipper2 (recall Sec.~\ref{subsec:dependencies}) and checking if the resulting area is at most $10^{-9}$ times the map area.
        \item \textit{Weak:} A special case where $\mathcal{A}$ and $\mathcal{R}$ return different non-Null outputs, but the query point is placed precisely on a weakly simple vertex of $\mathcal{W}$.
        A weakly simple vertex is one with more than one pair of incident edges.
        For example, two holes touching at a single vertex result in two pairs of incident edges for that vertex.
        \item \textit{Snap:} A special case where $\mathcal{A}$ and $\mathcal{R}$ return different non-Null outputs, but the query point has been snapped to some vertex of $\mathcal{W}$.
        \item \textit{Diff:} $\mathcal{A}$ and $\mathcal{R}$ return different non-Null outputs, and none of the special cases \textit{Weak} or \textit{Snap} apply.
    \end{itemize}
    To ensure compatibility with the above behaviors, all implementations have been modified to return Null when detecting that the query point lies outside $\mathcal{W}$.
    Assuming $\mathcal{R}$ is flawless, the most desirable behaviors are \textit{Same} and \textit{Null}, while the least desirable are certainly \textit{Crash} and \textit{Inf}, followed by \textit{Diff} and \textit{A0R1}.
    The special cases \textit{Weak} and \textit{Snap} are considered acceptable, as they are not necessarily indicative of an error but may hint at a feature-specific behavior.
    The same applies to \textit{A1R0} when the query point is $\epsilon_1$-close to an edge of $\mathcal{W}$ (as discussed in Sec.~\ref{subsec:robustness}), but in other cases, it is considered undesirable.
    Although the \textit{NoRef} case should be impossible assuming $\mathcal{R}$ is flawless, as we demonstrate later in this section, it had to be included as CGAL-TEA-CE occasionally encounters \textit{Crash} or \textit{Inf}.

    \subsection{Results}
    \label{subsec:results}

    \begin{table}
        \centering
        \tiny
        \setlength{\tabcolsep}{2pt}
        \caption{Runtime Behavior of the Evaluated Implementations}
        \label{tab:runtime_behavior}
        \begin{tabularx}{\columnwidth}{ll*{10}{>{\raggedleft\arraybackslash}X}}%
            \toprule
            Points         & $\mathcal{A}$ (tested imp.) & \textit{\%Crash} & \textit{\%Inf} & \textit{\%NoRef} & \textit{\%Null}  & \textit{\%A0R1}  & \textit{\%A1R0}  & \textit{\%Same}  & \textit{\%Weak}  & \textit{\%Snap}  & \textit{\%Diff}  \\
            \midrule
            \textit{In}    & CGAL-TEA-CE                 & -                & -              & -                & -               & -               & -               & 100.00          & -               & -               & -               \\
            & CGAL-TEA-CI                 & 7.418            & -              & -                & -               & -               & -               & 92.58           & -               & -               & -               \\
            & CGAL-RSA-CE                 & 5.391            & -              & -                & -               & -               & -               & 94.06           & -               & -               & 0.55            \\
            & CGAL-RSA-CI                 & 11.635           & -              & -                & -               & -               & -               & 87.92           & -               & -               & 0.44            \\
            & VisiLibity1                 & -                & -              & -                & -               & -               & -               & 82.86           & -               & -               & 17.14           \\
            & T\v{r}iVis                  & -                & -              & -                & -               & -               & -               & 100.00          & -               & -               & -               \\
            \midrule
            \textit{BB}    & CGAL-TEA-CE                 & -                & -              & -                & 42.66           & -               & -               & 57.34           & -               & -               & -               \\
            & CGAL-TEA-CI                 & 4.838            & -              & -                & 42.66           & -               & -               & 52.51           & -               & -               & -               \\
            & CGAL-RSA-CE                 & 3.456            & -              & -                & 42.66           & -               & -               & 53.61           & -               & -               & 0.28            \\
            & CGAL-RSA-CI                 & 7.471            & -              & -                & 42.66           & -               & -               & 49.64           & -               & -               & 0.23            \\
            & VisiLibity1                 & -                & -              & -                & 42.66           & -               & -               & 47.12           & -               & -               & 10.22           \\
            & T\v{r}iVis                  & -                & -              & -                & 42.66           & -               & -               & 57.34           & -               & -               & -               \\
            \midrule
            \textit{Ver}   & CGAL-TEA-CE                 & 0.012            & 0.018          & -                & 0.02            & -               & -               & 99.95           & -               & -               & -               \\
            & CGAL-TEA-CI                 & 4.282            & 0.029          & -                & 0.02            & -               & -               & 95.67           & -               & -               & -               \\
            & CGAL-RSA-CE                 & 3.868            & -              & 0.026            & 0.02            & -               & -               & 95.70           & -               & -               & 0.39            \\
            & CGAL-RSA-CI                 & 14.103           & -              & 0.024            & 0.02            & -               & -               & 64.74           & 0.80            & -               & 20.31           \\
            & VisiLibity1                 & -                & -              & 0.029            & -               & -               & 0.02            & 87.57           & 1.41            & -               & 10.97           \\
            & T\v{r}iVis                  & -                & -              & 0.029            & -               & -               & 0.02            & 98.48           & 1.47            & -               & -               \\
            \midrule
            \textit{NearV} & CGAL-TEA-CE                 & -                & 0.003          & -                & 39.45           & -               & -               & 60.54           & -               & -               & -               \\
            & CGAL-TEA-CI                 & 2.821            & 0.003          & -                & 39.45           & -               & -               & 57.72           & -               & -               & -               \\
            & CGAL-RSA-CE                 & 2.421            & -              & 0.003            & 39.45           & -               & -               & 57.64           & -               & -               & 0.49            \\
            & CGAL-RSA-CI                 & 5.506            & -              & 0.003            & 39.45           & -               & -               & 52.37           & 0.06            & -               & 2.60            \\
            & VisiLibity1                 & -                & -              & 0.003            & 32.79           & 0.0029          & 6.67            & 50.53           & 0.13            & 3.69            & 6.19            \\
            & T\v{r}iVis                  & -                & -              & 0.003            & 24.44           & -               & 15.01           & 58.01           & 0.13            & 2.40            & -               \\
            \midrule
            \textit{Mid}   & CGAL-TEA-CE                 & -                & -              & -                & 44.81           & -               & -               & 55.19           & -               & -               & -               \\
            & CGAL-TEA-CI                 & 2.971            & -              & -                & 44.81           & -               & -               & 52.21           & -               & -               & -               \\
            & CGAL-RSA-CE                 & 2.647            & -              & -                & 44.81           & -               & -               & 52.15           & -               & -               & 0.39            \\
            & CGAL-RSA-CI                 & 5.194            & -              & -                & 44.81           & -               & -               & 49.64           & -               & -               & 0.35            \\
            & VisiLibity1                 & -                & -              & -                & -               & -               & 44.81           & 47.47           & -               & 0.02            & 7.70            \\
            & T\v{r}iVis                  & -                & -              & -                & -               & -               & 44.81           & 55.19           & -               & -               & -               \\
            \midrule
            \textit{NearM} & CGAL-TEA-CE                 & -                & -              & -                & 25.84           & -               & -               & 74.16           & -               & -               & -               \\
            & CGAL-TEA-CI                 & 3.865            & -              & -                & 25.84           & -               & -               & 70.29           & -               & -               & -               \\
            & CGAL-RSA-CE                 & 3.268            & -              & -                & 25.84           & -               & -               & 70.36           & -               & -               & 0.53            \\
            & CGAL-RSA-CI                 & 6.538            & -              & -                & 25.84           & -               & -               & 67.14           & -               & -               & 0.48            \\
            & VisiLibity1                 & -                & -              & -                & 17.49           & -               & 8.35            & 64.54           & -               & 0.27            & 9.34            \\
            & T\v{r}iVis                  & -                & -              & -                & 13.11           & -               & 12.73           & 74.16           & -               & -               & -               \\
            \midrule
            Overall        & CGAL-TEA-CE                 & 0.002            & 0.003          & -                & 25.46           & -               & -               & 74.53           & -               & -               & -               \\
            & CGAL-TEA-CI                 & 4.366            & 0.005          & -                & 25.46           & -               & -               & 70.16           & -               & -               & -               \\
            & CGAL-RSA-CE                 & 3.508            & -              & 0.005            & 25.46           & -               & -               & 70.59           & -               & -               & 0.44            \\
            & CGAL-RSA-CI                 & 8.408            & -              & 0.004            & 25.46           & -               & -               & 61.91           & 0.14            & -               & 4.07            \\
            & VisiLibity1                 & -                & -              & 0.005            & 15.49           & 0.0005          & 9.98            & 63.35           & 0.26            & 0.66            & 10.26           \\
            & T\v{r}iVis                  & -                & -              & 0.005            & 13.37           & -               & 12.10           & 73.86           & 0.27            & 0.40            & -               \\
            \bottomrule
        \end{tabularx}
    \end{table}
    \begin{table}
        \centering
        \tiny
        \setlength{\tabcolsep}{1pt}
        \caption{Computational Time of the Evaluated Implementations}
        \label{tab:computational_time}
        \begin{tabularx}{\columnwidth}{lXrcrXrcrXrcrXr}%
            \toprule
            $\mathcal{A}$ (tested imp.) & & \multicolumn{3}{r}{Avg.\ init.\ [\textmu{}s]} & & \multicolumn{3}{r}{Avg.\ prep.\ [\textmu{}s]} & & \multicolumn{3}{r}{Avg.\ query [\textmu{}s]} & & PL [\%] \\
            \midrule
            CGAL-TEA-CE & & 820,793 & $\pm$ & 657,133 & & 8,129  & $\pm$ & 3,464 & & 142    & $\pm$ & 63     & & 51.1 \\
            CGAL-TEA-CI & & 618,622 & $\pm$ & 492,005 & & 5,887  & $\pm$ & 2,446 & & 88     & $\pm$ & 42     & & 64.8 \\
            CGAL-RSA-CE & & 818,563 & $\pm$ & 656,220 & &        & -     &       & & 8,861  & $\pm$ & 3,937  & & 0.8  \\
            CGAL-RSA-CI & & 619,976 & $\pm$ & 494,630 & &        & -     &       & & 5,243  & $\pm$ & 2,355  & & 1.1  \\
            VisiLibity1 & & 49      & $\pm$ & 20      & &        & -     &       & & 28,421 & $\pm$ & 12,912 & & 1.2  \\
            T\v{r}iVis  & & 1       & $\pm$ & 1       & & 16,303 & $\pm$ & 7,037 & & 9      & $\pm$ & 6      & & 2.7  \\
            \bottomrule
        \end{tabularx}
    \end{table}

    The runtime behavior of the evaluated implementations, presented as a percentage of the number of query points is shown in Tab.~\ref{tab:runtime_behavior}.
    For better readability, zero values have been replaced with a dash (-).
    Recall that CGAL-TEA-CE serves as the reference implementation $\mathcal{R}$; thus, where $\mathcal{R}$ defines the behavior, the CGAL-TEA-CE values represent a form of `comparison to itself.'

    First, note that CGAL implementations have a high crash rate, making them largely unusable.
    The exception is the reference implementation, CGAL-TEA-CE, which crashed only four times and looped seven times out of 204,000 runs.
    All these issues occurred with the \textit{Ver} or \textit{NearV} sets.
    In contrast, VisiLibity1 and T\v{r}iVis showed no crashes or infinite loops.
    Additionally, CGAL implementations, including those with exact constructions (CE), are not mutually consistent in their outputs, as shown by the 0.44\% occurrence of \textit{Diff} for CGAL-RSA-CE\@.
    Our manual inspections of some of these cases revealed that CGAL-TEA-CE was the one providing correct outputs, making it the most reliable among the CGAL options.
    VisiLibity1, while not crashing or looping, is inconsistent with the reference in 10.26\% of cases, as indicated by \textit{Diff}, and is therefore unreliable as well.

    Moving our attention to T\v{r}iVis, we see that it is the most reliable, as it never crashed or looped and is consistent with the reference outputs in all cases, except for the \textit{Weak}, \textit{Snap}, and \textit{A1R0} cases, as evidenced by zero occurrences of \textit{Diff} and \textit{A0R1}\@.
    As previously mentioned, the \textit{Weak}, \textit{Snap}, and \textit{A1R0} cases are not necessarily indicative of an error but may hint at feature-specific behavior.
    The \textit{Weak} case occurs only for the \textit{Ver} and \textit{NearV} sets and suggests a different strategy used in handling weakly-simple vertex queries in T\v{r}iVis compared to the reference.
    Upon manual inspection, we found that CGAL-TEA-CE selects only one pair of incident edges as the initial edges for expansion, while T\v{r}iVis considers all of them, making it consistent with our visibility region definition.
    The \textit{Snap} case occurs only for the \textit{NearV} set and is caused by the vertex snapping technique, as described in Sec.~\ref{subsec:structure}, which helps T\v{r}iVis compute all intersection points correctly and prevent more significant errors.
    Upon manual inspection, we found that T\v{r}iVis's outputs for the \textit{Snap} cases align with our expectations and show no indications of more significant errors.
    The \textit{A1R0} case occurs only for the \textit{Mid} and \textit{NearM} sets.
    For the \textit{Mid} set, T\v{r}iVis returns non-null outputs in 100\% of cases.
    This suggests that T\v{r}iVis can compute more visibility polygons in cases where the query point is $\epsilon_1$-close to an edge of $\mathcal{W}$, as discussed in Sec.~\ref{subsec:robustness}, while the reference implementation discards some of these query points as being outside $\mathcal{W}$.

    The computational time of the evaluated implementations is presented in Tab.~\ref{tab:computational_time}.
    This table shows, for the \textit{In} query set, the average initialization (init.), preprocessing (prep.), and query times, along with the percentage of time (out of the query time) spent on the point location query (PL).
    Note that init.\ and prep.\ times are averaged over the 34 maps, while the query time is averaged over the 34,000 query points.
    The init.\ time represents the duration needed to construct the implementation-specific representation of the environment (e.g., \textsc{CGAL::Arrangement\_2} for CGAL) from a simple polygonal representation.
    The prep.\ time covers the more formal preprocessing operations.
    While RSA and VisiLibity1 do not require preprocessing, the prep.\ time for the TEA implementations includes the time needed to construct the triangular mesh and bucketing structures for the point location query in T\v{r}iVis.
    Notably, T\v{r}iVis requires twice as much prep.\ time compared to CGAL-TEA-CE, but the total pre-query time is almost two orders of magnitude lower due to the high init.\ times of the CGAL implementations.
    Moreover, T\v{r}iVis has the lowest query time among all evaluated implementations.
    It is one order of magnitude lower than CGAL’s TEA implementations, four orders lower than CGAL’s RSA implementations, and five orders lower than VisiLibity1.
    It is noteworthy that CGAL’s TEA implementations spend about half of the query time on the point location query, which could potentially be improved with a more efficient point location algorithm.%
    \footnote{
        It is important to note that all CGAL visibility polygon implementations assume that the query point is located inside $\mathcal{W}$ and require the user to provide the edge or vertex on which the query point is located, if applicable.
        This necessitates using the point location implementations from the CGAL 2D Arrangements package prior to performing the visibility query.
        We use the \textsc{CGAL::Arr\_\allowbreak{}walk\_\allowbreak{}along\_\allowbreak{}line\_\allowbreak{}point\_\allowbreak{}location} implementation, which was the fastest among those available in CGAL that provided correct results without crashing, based on our preliminary experiments.
    }
    Even if we halve the average query time of CGAL-TEA-CE, T\v{r}iVis still outperforms it by nearly a factor of eight.

    \section{Concluding Remarks}
    \label{sec:conclusion}

    This paper presented T\v{r}iVis, a C++ library for computing various visibility-related queries in polygonal environments.
    We have demonstrated that T\v{r}iVis excels in several aspects compared to similar available implementations:
    \begin{itemize}
        \item \emph{Versatility:} T\v{r}iVis offers an extensive set of features, each with a specialized, user-friendly interface.
        These features include computing visibility polygons, performing two-points and ray-shooting visibility queries, identifying visible points and vertices, constructing visibility graphs, and executing rapid point location queries.
        It also provides various utility functions for managing polygonal environments and query outputs.
        Additionally, T\v{r}iVis supports efficient computations of all queries with an optional limited range constraint.
        \item \emph{Reliability:} T\v{r}iVis is characterized by its reliable and predictable behavior.
        It avoids crashing or infinite looping and consistently produces outputs that align with user expectations, as confirmed by our evaluation on highly complex query instances.
        \item \emph{Performance:} With an average query time of 9\,\textpm\,6\,\textmu{}s, T\v{r}iVis outperforms all other implementations by at least an order of magnitude, while maintaining preprocessing times below 20\,ms for the benchmark instances.
    \end{itemize}
    Moreover, the core functionality is independent of external libraries.
    T\v{r}iVis is freely available for private, research, and institutional use at \url{https://github.com/janmikulacz/trivis}.

    \bibliographystyle{IEEEtran}
    \bibliography{main}
    \pagebreak

\end{document}